\documentclass[letterpaper, 10 pt, conference]{ieeeconf}
\usepackage{fix-cm}
\usepackage{amsmath}
\usepackage{amssymb}
\usepackage{physics}
\usepackage{graphicx}
\usepackage{url}
\usepackage{booktabs}
\usepackage{subcaption}
\usepackage{multirow}
\usepackage{pgfplots}
\usepackage[compatibility=false]{caption}
\usepackage{tabularx}
\usepgfplotslibrary{groupplots}
\pgfplotsset{compat=1.18}
\usepackage[font=footnotesize,labelfont=bf]{caption}
\usepackage[colorlinks=true,
            linkcolor=blue,
            citecolor=blue,
            urlcolor=blue]{hyperref}

\IEEEoverridecommandlockouts
\title{\LARGE \bf
WaveLander: A Generalizable Hierarchical Control Framework for UAV Landing on Wave-Disturbed Platforms via Reinforcement Learning\vspace{-0.8em}
}

\author{
Chun-Kit Li$^{*,1,4}$, Iok Long Sit$^{*,2}$, Ming Fung Siu$^{3}$,
Ka Yu Kui$^{3,4}$, Hin Wang Lin$^{3,4}$,\\
Pengyu Wang$^{\dag,3,4}$ and Ling Shi$^{\dag,3,4}$, \textit{Fellow, IEEE}
\\[0.9em]
\makebox[\textwidth][c]{%
\begin{minipage}{1\textwidth}
    \centering
    \includegraphics[
        width=\linewidth,
        keepaspectratio
    ]{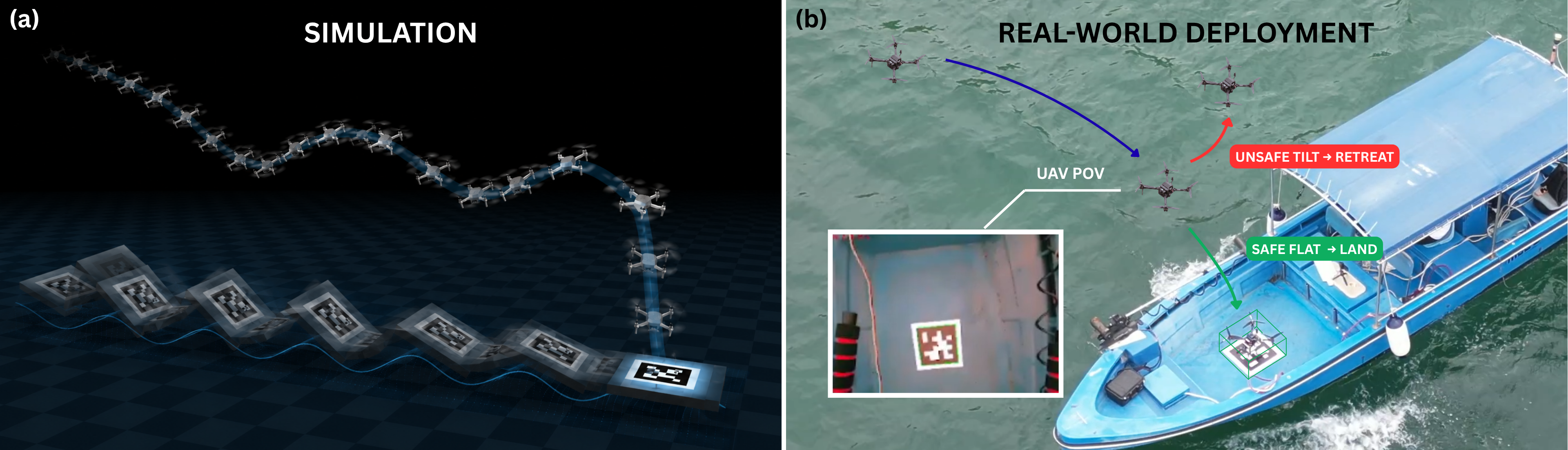}\\[-0.3em]
    \refstepcounter{figure}\label{fig:title_overview}
    {\small \textbf{Fig.~\thefigure.} Overview visualization of WaveLander. (a) MuJoCo simulation rendering of timing-aware landing on a wave-disturbed platform. (b) Real-world deployment illustration showing representative landing and retreat behaviors.}
\end{minipage}
}
\thanks{*The first two authors contributed equally.}%
\thanks{$\dag$Corresponding authors: Ling Shi and Pengyu Wang.}%
\thanks{$^{1}$Chun-Kit Li is with the Department of Mathematics, The Hong Kong University of Science and Technology, Hong Kong SAR. {\tt\small ckliaq@connect.ust.hk}}%
\thanks{$^{2}$Iok Long Sit is with the Division of Integrative Systems and Design, The Hong Kong University of Science and Technology, Hong Kong SAR. {\tt\small ilsit@connect.ust.hk}}%
\thanks{$^{3}$Ming Fung Siu, Ka Yu Kui, Hin Wang Lin, Pengyu Wang and Ling Shi are with the Department of Electronic and Computer Engineering, The Hong Kong University of Science and Technology, Hong Kong SAR. {\tt\small \{mfsiu, kykui, hwlinaa, pwangat\}@connect.ust.hk, eesling@ust.hk}}%
\thanks{$^{4}$Chun-Kit Li, Ka Yu Kui, Hin Wang Lin, Pengyu Wang and Ling Shi are also with the Cheng Kar-Shun Robotics Institute, The Hong Kong University of Science and Technology, Hong Kong SAR. In addition, Ling Shi is with the Department of Chemical and Biological Engineering, The Hong Kong University of Science and Technology, Hong Kong SAR.}%
}

\begin{document}

\maketitle

\setcounter{figure}{1}

\begin{abstract}
Autonomous landing of unmanned aerial vehicles (UAVs) on wave-disturbed marine platforms remains challenging due to stochastic platform motion, time-varying platform attitude, and uncertain touchdown conditions. Existing model-based methods often require accurate motion prediction and online optimization, while end-to-end learning approaches may suffer from high training complexity and limited interpretability. This paper presents WaveLander, a hierarchical control framework via reinforcement learning (RL) that decouples vertical landing decision-making from low-level flight stabilization. The RL policy maps a compact platform-relative observation to a scalar vertical velocity reference, while a conventional low-level flight controller maintains attitude stability and lateral tracking. This formulation reduces dynamic platform landing to a low-dimensional, timing-aware control problem and enables smooth landing behavior without explicit switching rules. Simulation results under randomized wave-induced platform motions show that WaveLander achieves robust landing performance and generalizes to unseen disturbance conditions, demonstrating the potential of hierarchical learning-based control for marine UAV recovery.
\end{abstract}

\section{Introduction}

Unmanned aerial vehicles and unmanned surface vehicles (USVs) are increasingly used together in marine operations. In these heterogeneous robotic systems, UAVs provide fast aerial sensing, wide-area surveillance, and target localization, while USVs offer longer endurance, greater payload capacity, and direct interaction with the water surface. This cooperation is especially valuable in tasks such as search and rescue at sea, offshore inspection, environmental monitoring, and pollution response, where aerial observation and surface-level action must work together with limited human intervention~\cite{lin2024coastal, zhang2025aerial, li2024separation}. In such settings, the ability of a UAV to reliably land on a ship or USV is crucial, since it determines whether the vehicle can be safely recovered, recharged, redeployed, and sustained in long-duration missions.

Despite its practical importance, autonomous UAV landing on wave-disturbed marine platforms remains difficult. Unlike fixed infrastructure~\cite{wang2022quadrotor}, ship and USV decks are affected by wave-induced platform motion, producing time-varying heave, roll, and pitch motions. These disturbances create major uncertainty in the relative pose and velocity between the UAV and the landing platform, particularly near touchdown, where small errors may cause large impact forces or instability. Wind, sensor noise, and imperfect state estimation further increase the challenge. Therefore, UAV landing in ocean environments is not simply a trajectory tracking problem, but a coupled decision-and-control task that demands precise timing, robustness, and adaptability.

Recent studies have investigated model-based and
optimization-driven solutions for UAV landing on marine platforms.
For example, Gupta et al.~\cite{gupta2023landing} proposed an MPC-based
framework that predicts vessel motion online and performs landing during
favorable near-zero-tilt windows. Such methods demonstrate the importance
of exploiting the temporal structure of wave-induced platform motion.
However, they typically require online motion prediction, carefully
designed objective functions, and repeated optimization during flight.
Their performance may also depend on the accuracy of the predicted
platform motion, which can be degraded by stochastic disturbances,
sensor noise, and short observation windows.

Reinforcement learning offers an alternative paradigm by directly learning
a feedback policy through interaction, avoiding explicit prediction
of future platform motion and repeated online optimization. This is particularly
appealing for wave-disturbed landing, where the core challenge lies in making
timing-sensitive decisions under uncertain
platform motion. However, existing RL-based approaches~\cite{polvara2018toward}
typically rely on high-dimensional visual inputs and discrete action policies, leading to high training complexity, coarse control,
and limited integration with reliable low-level flight controllers, which
restricts their applicability in safety-critical dynamic scenarios.

To address these limitations, we propose \textit{WaveLander}, a hierarchical
control framework via RL that decouples vertical landing decision-making
from low-level flight stabilization. A conventional low-level flight controller ensures
robust tracking, while a learned policy outputs a continuous vertical velocity reference
relative to the moving platform. This compact formulation reduces learning
complexity, improves interpretability, and enables smooth landing behavior
without heuristic switching. The main contributions of this paper are summarized as follows:
\begin{itemize}
    \item A hierarchical control framework via RL, termed WaveLander, is proposed for UAV
    landing on wave-disturbed platforms by decoupling high-level vertical
    landing decisions from low-level flight stabilization.

    \item A compact platform-relative policy interface maps relative height and
    platform attitude to a continuous vertical velocity reference, reducing
    learning complexity and simplifying deployment.

    \item An attitude-aware reward-shaping design is introduced to encourage
    descent, hold, and retreat behaviors, enabling the learned policy to
    produce continuous vertical commands without threshold-based switching
    during deployment.

    \item The framework is evaluated through randomized simulation, SITL transfer, and a representative deployment-oriented real-world test under dynamic platform motion.
\end{itemize}

\section{Related Work}

Autonomous UAV landing on moving platforms, especially shipborne and USV decks, has been widely studied using control-theoretic, perception-driven, and learning-based approaches. Existing methods can be broadly grouped into two categories: model-based control methods and learning-based methods.

\subsection{Model-based Control Methods}

Traditional approaches typically formulate the landing problem as trajectory tracking, state estimation, or optimal control, relying on explicit models of UAV dynamics and platform motion, often with prior prediction of platform movement. For example, Gupta \emph{et al.}~\cite{gupta2023landing} proposed an MPC-based framework that predicts vessel motion online and schedules landing during favorable time windows. To improve robustness under wave disturbances, Li \emph{et al.}~\cite{li2022synchronized} introduced a vision-guided synchronized motion strategy combining Bi-LSTM-based prediction with PID control, while Li \emph{et al.}~\cite{li2023nmpc} further developed an NMPC-based cooperative tracking and landing framework to exploit favorable heave phases for precise touchdown. In addition, Xu \emph{et al.}~\cite{xu2024manipulator} proposed a manipulator-assisted system with a robust NMPC controller to actively capture UAVs under disturbances. Despite their interpretability and safety guarantees, these model-based methods rely heavily on accurate modeling and prediction, which become unreliable in realistic maritime environments with stochastic wave-induced motion, leading to degraded performance under highly dynamic conditions.

\subsection{Learning-based Methods}

To improve adaptability under uncertain disturbances, reinforcement learning has been increasingly explored for autonomous landing. Early work by Polvara \emph{et al.}~\cite{polvara2018toward} demonstrated end-to-end UAV landing using a hierarchical Deep Q-Network framework directly from low-resolution visual inputs, highlighting the potential of learning-based policies to generalize across diverse conditions. Building on this direction, Rodriguez-Ramos \emph{et al.}~\cite{rodriguez2019deep} showed that deep reinforcement learning can learn continuous landing control policies for moving platforms and achieve improved robustness compared with conventional low-level flight controllers. More recent studies have further extended learning-based approaches to dynamic landing scenarios with improved adaptability and success rates~\cite{goldschmid2024rl}. In maritime environments, such methods have also been applied to vision-based ship landing and adaptive landing on highly dynamic platforms~\cite{saj2022robust,peter2024landerai}, demonstrating the ability to capture timing-aware and disturbance-tolerant behaviors directly through interaction, without relying on manually designed landing logic. However, many existing approaches attempt to learn full end-to-end control policies, which increases training complexity and limits interpretability. In contrast, this work adopts a hierarchical formulation in which the RL policy focuses on high-level vertical landing decisions, while a conventional low-level flight controller ensures stabilization and tracking.

\label{subsec:system_overview}

\begin{figure*}[!t]
    \centering
    \includegraphics[width=\textwidth]{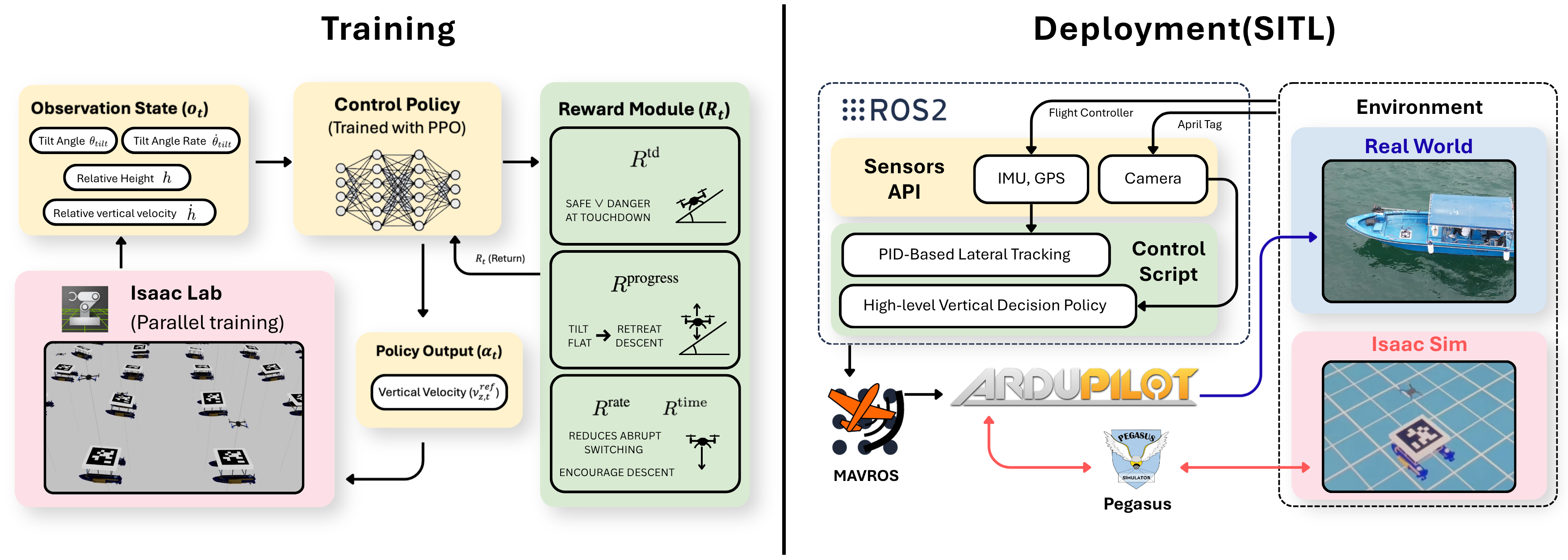}
    \caption{System overview of the proposed framework.}
    \label{fig:overview}
\end{figure*}

\section{Methodology}
\label{sec:methodology}

This section presents WaveLander, a hierarchical control framework via RL for UAV landing on wave-disturbed platforms. The learned component outputs a scalar vertical
velocity reference from a compact platform-relative observation, while a
conventional low-level flight controller handles stabilization, velocity
tracking, and lateral alignment. Timing-related landing behavior is encouraged
through reward shaping rather than explicit platform-motion prediction or
threshold-based switching.

\subsection{System Overview}

As shown in Fig.~\ref{fig:overview}, WaveLander adopts a hierarchical control
structure. The high-level policy generates a scalar vertical velocity reference
\(v^{\mathrm{ref}}_{z,t}\), while the low-level flight controller tracks the
commanded motion and maintains attitude and lateral stability. This separation
allows learning to focus on high-level vertical landing decisions while
preserving compatibility with standard UAV flight-control stacks.

\subsection{Problem Formulation}
\label{subsec:problem_formulation}

The landing task is formulated as a high-level decision-making problem
rather than a direct motor-control problem and is modeled as an episodic
discounted Markov decision process (MDP)~\cite{sutton2018reinforcement},
\begin{equation}
\mathcal{M}=(\mathcal{S},\mathcal{A},\mathcal{P},\mathcal{R},\gamma),
\end{equation}
where $\mathcal{S}$ denotes the underlying landing-related state space,
$\mathcal{A}$ is the high-level action space, $\mathcal{P}$ represents the
closed-loop transition dynamics induced by the UAV, the low-level controller,
and the wave-disturbed platform, $\mathcal{R}$ is the reward function, and
$\gamma\in[0,1)$ is the discount factor.

In practice, the actor does not receive the full state $s_t\in\mathcal{S}$.
Instead, it is conditioned on a compact platform-relative observation $o_t$
for deployment simplicity. At each time step, the policy outputs a scalar
high-level action,
\begin{equation}
a_t=\pi_\psi(o_t),
\end{equation}
which is mapped to a vertical velocity reference and tracked by the low-level
flight controller. The objective is to learn policy parameters $\psi$ that
maximize the expected discounted return:
\begin{equation}
\psi^* = \operatorname*{arg\,max}_{\psi} \mathbb{E}_{\pi_\psi} \left[ \sum_{t=0}^{\infty} \gamma^t r_t \right]
\end{equation}

The optimal high-level policy is then given by $\pi_{\psi^*}$.

\subsection{Wave-Disturbed Platform Model}
\label{subsec:wave_platform}

To train and evaluate the policy under marine-like motion, the landing platform
is modeled as a kinematically driven rigid body with heave, roll, and pitch
disturbances. Since the focus is the landing policy rather than high-fidelity
hydrodynamics, a lightweight randomized wave generator is used. The disturbance
along each motion axis is generated by a superposition of sinusoidal components:
\begin{equation}
d_{\mathrm{axis}}(t)
=
\sum_{i=1}^{K}
A_i^{(\mathrm{axis})}
\sin
\left(
\omega_i t+\varphi_i^{(\mathrm{axis})}
\right),
\label{eq:wave_multisine}
\end{equation}
where \(K\) is the number of components, and
\(A_i^{(\mathrm{axis})}\), \(\omega_i\), and
\(\varphi_i^{(\mathrm{axis})}\) are the amplitude, angular frequency, and phase
of the \(i\)-th component. The frequencies are shared across heave, roll, and
pitch, while amplitudes and phases are sampled independently to generate
varying platform tilt directions.

To improve generalization, the wave parameters are resampled at the beginning
of each episode, with larger amplitudes assigned to lower-frequency components.
Temporally correlated stochastic variation is added using an
Ornstein--Uhlenbeck process~\cite{uhlenbeck1930theory}:
\begin{equation}
n_{t+1}
=
\rho n_t+\sigma A_{\max}\epsilon_t,
\qquad
\epsilon_t\sim\mathcal{N}(0,1),
\label{eq:ou_noise}
\end{equation}
where \(\rho\) controls temporal correlation, \(\sigma\) controls the noise
strength, and \(A_{\max}\) is the maximum disturbance amplitude. To avoid
abrupt platform motion, the resulting signal is smoothed before being applied:
\begin{equation}
s_{t+1}
=
(1-\alpha)s_t
+
\alpha
\left(
d_{\mathrm{axis}}(t)+n_t
\right),
\label{eq:wave_smoothing}
\end{equation}
where \(\alpha\) is the smoothing coefficient. The smoothed signals are applied
to heave, roll, and pitch, while yaw is fixed.

\subsection{Platform-Relative Perception and Control}
\label{subsec:perception_lateral}

For deployment, the platform-relative state is obtained from an AprilTag
mounted on the landing platform. The policy does not use raw images; instead,
the detector provides a structured relative pose measurement. Let
\(P^{\mathrm{cam}}_{\mathrm{tag}}\) denote the detected tag position in the
camera frame. With calibrated camera-to-body extrinsics, the tag position is
first expressed in the UAV body frame:
\begin{equation}
P^{\mathrm{body}}_{\mathrm{body}\rightarrow\mathrm{tag}}
=
R^{\mathrm{body}}_{\mathrm{cam}}P^{\mathrm{cam}}_{\mathrm{tag}}
+
P^{\mathrm{body}}_{\mathrm{body}\rightarrow\mathrm{cam}} .
\label{eq:tag_body_transform}
\end{equation}
Let \(p^{\mathrm{body}}_{\mathrm{body}\rightarrow\mathrm{gear}}\) be the fixed
vector from the UAV body-frame origin to the lowest landing-gear point. The
body-frame vector from the landing gear to the pad reference point is
\begin{equation}
P^{\mathrm{body}}_{\mathrm{gear}\rightarrow\mathrm{pad}}
=
P^{\mathrm{body}}_{\mathrm{body}\rightarrow\mathrm{tag}}
-
p^{\mathrm{body}}_{\mathrm{body}\rightarrow\mathrm{gear}} .
\label{eq:gear_to_pad_body}
\end{equation}
This vector is then expressed in the gravity-aligned world frame:
\begin{equation}
P^{\mathrm{world}}_{\mathrm{gear}\rightarrow\mathrm{pad}}
=
R^{\mathrm{world}}_{\mathrm{body}}
P^{\mathrm{body}}_{\mathrm{gear}\rightarrow\mathrm{pad}} .
\label{eq:gear_to_pad_world}
\end{equation}
The vertical height used by the policy is
\begin{equation}
h
=
-e_z^\top
P^{\mathrm{world}}_{\mathrm{gear}\rightarrow\mathrm{pad}},
\label{eq:relative_height}
\end{equation}
where \(e_z=[0,0,1]^\top\). Thus, \(h_t=0\) corresponds to nominal contact
between the landing gear and the pad surface.

The platform position error is defined as
\begin{equation}
\mathbf{e}_{xy} = \mathbf{p}^{\mathrm{tag}}_{xy} - \mathbf{p}_{xy}.
\end{equation}
Lateral motion is handled by an SE(2) PID tracking controller:
\begin{equation}
\mathbf{v}^{\mathrm{ref}}_{xy} = K_p^{xy} \mathbf{e}_{xy} + K_i^{xy} \int \mathbf{e}_{xy} \, dt + K_d^{xy} \dot{\mathbf{e}}_{xy}.
\end{equation}
For vertical motion, the desired vertical velocity \(v^{\mathrm{ref}}_z\) is directly provided by the RL policy as the action \(a_t = \pi_\psi(o_t)\). The overall desired velocity vector is \(\mathbf{v}^{\mathrm{ref}} = [\mathbf{v}^{\mathrm{ref}}_{xy},\; v^{\mathrm{ref}}_z]^\top\).

Define the velocity error vector as
\begin{equation}
\mathbf{e}_v = \mathbf{v}^{\mathrm{ref}} - \mathbf{v}.
\end{equation}
A low-level PID controller with gravity compensation then computes the desired acceleration:
\begin{equation}
\mathbf{a}^{\mathrm{des}} = K_p^{v} \mathbf{e}_v + K_i^{v} \int \mathbf{e}_v \, dt + K_d^{v} \dot{\mathbf{e}}_v + [0,\;0,\;g_0]^\top.
\end{equation}
The desired thrust direction is obtained by normalizing \(\mathbf{a}^{\mathrm{des}}\):
\begin{equation}
\mathbf{z}_{\mathrm{des}} = \frac{\mathbf{a}^{\mathrm{des}}}{\|\mathbf{a}^{\mathrm{des}}\|}.
\end{equation}
The desired thrust direction \(\mathbf{z}_{\mathrm{des}}\) is then sent to a low-level SE(3) attitude controller, which computes the required body torques and collective thrust.

\subsection{Observation and Action Design}
\label{subsec}

The policy uses a compact platform-relative observation to capture the vertical
landing state and the attitude trend of the moving platform. Let $h_t$ and
$\dot h_t$ denote the relative height and vertical velocity.

The platform attitude is represented by the quaternion
$q=(q_{w},q_{x},q_{y},q_{z})$. The roll and pitch angles are computed
as
\begin{equation}
\phi =
\operatorname{atan2}
\left(
2(q_{w}q_{x}+q_{y}q_{z}),
1-2(q_{x}^2+q_{y}^2)
\right),
\label{eq:roll_angle}
\end{equation}
\begin{equation}
\theta =
\arcsin
\left(
\operatorname{clip}
\left(
2(q_{w}q_{y}-q_{z}q_{x}),-1,1
\right)
\right).
\label{eq:pitch_angle}
\end{equation}

The platform tilt magnitude and its time derivative are defined as
\begin{equation}
\theta_{tilt} = \sqrt{\phi^2 + \theta^2}, \qquad
\dot{\theta}_{tilt} = \frac{d\theta_{\text{tilt}}}{dt}.
\label{eq:tilt}
\end{equation}

The policy observation is
\begin{equation}
o=[h,\dot h,\theta_{tilt},\dot{\theta}_{tilt}]^\top
\in \mathbb{R}^{4}.
\label{eq}
\end{equation}

The deployed vertical command is obtained by clipping the raw policy output:
\begin{equation}
v^{\mathrm{ref}}_{z}
=
\operatorname{clip}\!\left(
a, -0.45, 0.45
\right)\ \mathrm{m/s}.
\label{eq:action_clip}
\end{equation}

\subsection{Reward Design}
\label{subsec:reward_design}

The reward is designed to shape the high-level vertical decision under
wave-induced platform motion. The objective is to encourage descent when the
platform condition is favorable while discouraging further descent and
promoting upward retreat when the UAV is close to an excessively tilted
platform. This behavior is induced without introducing a threshold-based
switching policy; instead, the landing condition is represented through
continuous height- and attitude-dependent gates.

At each control step, the total reward is defined as
\begin{equation}
R_t =
R_t^{\mathrm{prog}}
+
R_t^{\mathrm{td}}
+
R_t^{\mathrm{rate}}
+
R_t^{\mathrm{time}},
\label{eq:total_reward}
\end{equation}
where the four terms correspond to vertical progress modulation, touchdown
quality, command smoothness, and delay suppression, respectively.

\subsubsection{Proximity and Attitude Assessment}

The landing condition is evaluated using two continuous gates. The height gate
measures the proximity to the platform, while the attitude gate measures the
suitability of the platform orientation:
\begin{equation}
\begin{aligned}
g_h &=
\exp\!\left(-\frac{h_t}{\sigma_h}\right), \\
g_{\theta} &=
\exp\!\left(
-\frac{\max(0,\,\theta_{\mathrm{tilt},t}-\theta_{\mathrm{safe}})}
{\sigma_\theta}
\right).
\end{aligned}
\label{eq:gates}
\end{equation}
Here, \(h_t\) is the relative height, \(\theta_{\mathrm{tilt},t}\) is the
platform tilt magnitude, and \(\theta_{\mathrm{safe}}\) is a soft attitude
level used only for reward shaping. It is not the hard success threshold used
in evaluation. The parameters \(\sigma_h\) and \(\sigma_\theta\) determine the
sensitivity to height and tilt, respectively. Therefore, \(g_h\) increases as
the UAV approaches the platform, whereas \(g_\theta\) decreases smoothly when
the platform tilt exceeds the desired landing range.

\subsubsection{Descent--Retreat Modulation}

The two gates are combined into an adaptive coefficient that determines whether
vertical motion is rewarded as descent or retreat:
\begin{equation}
\alpha_t
=
1 - 2g_h(1-g_\theta).
\label{eq:alpha}
\end{equation}
The progress term is defined as
\begin{equation}
R_t^{\mathrm{prog}}
=
-w_{\mathrm{prog}}\,\dot h_t\,\alpha_t ,
\label{eq:progress_reward}
\end{equation}
where \(\dot h_t<0\) denotes descent and \(\dot h_t>0\) denotes upward retreat.
When the UAV is far from the platform, \(g_h\approx 0\), so
\(\alpha_t\approx 1\) and descent is encouraged. When the UAV is close to a
level platform, \(g_h\approx 1\) and \(g_\theta\approx 1\), resulting again in
\(\alpha_t\approx 1\), which encourages continued descent. In contrast, when
the UAV is close to a highly tilted platform, \(g_h\approx 1\) and
\(g_\theta\approx 0\), giving \(\alpha_t\approx -1\). In this case, further
descent is penalized and upward retreat is rewarded. Intermediate cases are
handled continuously through the same coefficient.

\subsubsection{Touchdown Quality Evaluation}

Touchdown is rewarded according to both platform attitude and vertical impact
speed. Let \(C_t\in\{0,1\}\) denote the contact indicator. The touchdown quality
term is given by
\begin{equation}
R_t^{\mathrm{td}}
=
C_t w_{\mathrm{td}}
\exp\!\left(
-\frac{\theta_{\mathrm{tilt},t}^2}
{2\sigma_{td,\theta}^2}
\right)
\exp\!\left(
-\frac{\dot h_t^2}
{2\sigma_{td,\dot h}^2}
\right),
\label{eq:terminal_reward}
\end{equation}
where \(\sigma_{td,\theta}\) and \(\sigma_{td,\dot h}\) specify the shaping
scales for touchdown tilt and vertical speed, respectively. This term assigns
the highest reward to near-level and low-speed touchdowns, while providing a
graded quality measure for imperfect contacts. The final success rate reported
in evaluation is determined separately by the touchdown criteria, rather than
by the soft shaping level \(\theta_{\mathrm{safe}}\).

\subsubsection{Command Smoothness and Delay Suppression}

To avoid oscillatory or aggressive command changes, an action-rate penalty is
included:
\begin{equation}
R_t^{\mathrm{rate}}
=
-w_a(a_t-a_{t-1})^2 .
\label{eq:rate}
\end{equation}
In addition, a small time penalty is applied before touchdown:
\begin{equation}
R_t^{\mathrm{time}}
=
-w_{\mathrm{time}}(1-C_t).
\label{eq:time}
\end{equation}
This term discourages excessive waiting while keeping the dominant objective on
attitude-aware landing quality.

\begin{table}[htbp]
\centering
\caption{Reward parameters.}
\label{tab:reward_params}
\footnotesize
\renewcommand{\arraystretch}{1.08}
\setlength{\tabcolsep}{3.6pt}
\begin{tabular}{@{}llcl@{}}
\hline
\textbf{Group} & \textbf{Parameter} & \textbf{Symbol} & \textbf{Value} \\
\hline
\multirow{3}{*}{\textit{Condition}}
& Height scale & \(\sigma_h\) & \(1.1~\mathrm{m}\) \\
& Soft safe tilt & \(\theta_{\mathrm{safe}}\) & \(0.08~\mathrm{rad}\) \\
& Tilt sensitivity & \(\sigma_\theta\) & \(0.012~\mathrm{rad}\) \\
\hline
\multirow{2}{*}{\textit{Touchdown}}
& Tilt shaping scale & \(\sigma_{td,\theta}\) & \(0.10~\mathrm{rad}\) \\
& Speed shaping scale & \(\sigma_{td,\dot h}\) & \(0.45~\mathrm{m/s}\) \\
\hline
\multirow{4}{*}{\textit{Weights}}
& Progress & \(w_{\mathrm{prog}}\) & \(10\) \\
& Touchdown & \(w_{\mathrm{td}}\) & \(1000\) \\
& Action-rate & \(w_a\) & \(0.1\) \\
& Time penalty & \(w_{\mathrm{time}}\) & \(0.01\) \\
\hline
\end{tabular}
\end{table}

\subsection{Training Setup}
\label{subsec:training_setup}

The policy is trained in NVIDIA Isaac Lab~\cite{mittal2025isaaclab} using
Proximal Policy Optimization (PPO)~\cite{schulman2017ppo} with the
RSL-RL implementation~\cite{schwarke2025rslrl}. Running mean--standard
deviation normalization is applied to the policy inputs, and the actor
normalizer is exported together with the deployed JIT policy. The actor and
critic are implemented as GRU-based recurrent networks with ELU activations.
The actor uses hidden layers of \([128,64]\), while the critic uses
\([256,128]\); both use a GRU hidden dimension of \(64\). The initial action
noise standard deviation is set to \(1.5\).

The UAV is modeled as a single prismatic joint constrained to vertical
translation. A low-level velocity controller tracks the commanded vertical
velocity, with randomized damping gains to capture variability in the
closed-loop vertical response. Additional domain randomization is applied during training,
including a random action delay of \(0\)--\(3\) control steps, a
\(\pm30\%\) mass perturbation, and resampled wave parameters in each episode,
covering the number of sinusoidal components, frequency range, and
Ornstein--Uhlenbeck noise intensity.

A lightweight curriculum on the platform tilt magnitude is used to improve
training stability. The policy is first trained under smaller pad tilt angles
and is then progressively exposed to larger tilt disturbances. The final
checkpoint is selected according to validation performance under randomized
tilt conditions.

To improve robustness against state-estimation errors, Gaussian observation
noise is added to the actor inputs but not to the critic. The relative position
and vertical velocity observations receive zero-mean noise with standard
deviation \(0.02\), while the pad attitude features receive zero-mean noise
with standard deviation \(0.01\).

For PPO, each iteration collects \(64\) steps from \(30000\) parallel
environments, and training is conducted for \(1000\) iterations. The learning
rate is initialized at \(1\times10^{-3}\) and adjusted using an adaptive KL
schedule with a desired KL divergence of \(0.01\). Each update uses \(5\)
epochs and \(16\) mini-batches per epoch. The discount factor and GAE parameter
are set to \(\gamma=0.995\) and \(\lambda=0.97\), respectively. The PPO clip
range is \(0.2\), the entropy coefficient is \(0.008\), the value-loss
coefficient is \(1.0\), clipped value loss is enabled, and the maximum gradient
norm is limited to \(1.0\). Training is conducted on a workstation with an AMD
Threadripper PRO 5965WX CPU and an NVIDIA RTX A5000 GPU.

\begin{figure*}[t]
    \centering
    \resizebox{\textwidth}{!}{\input{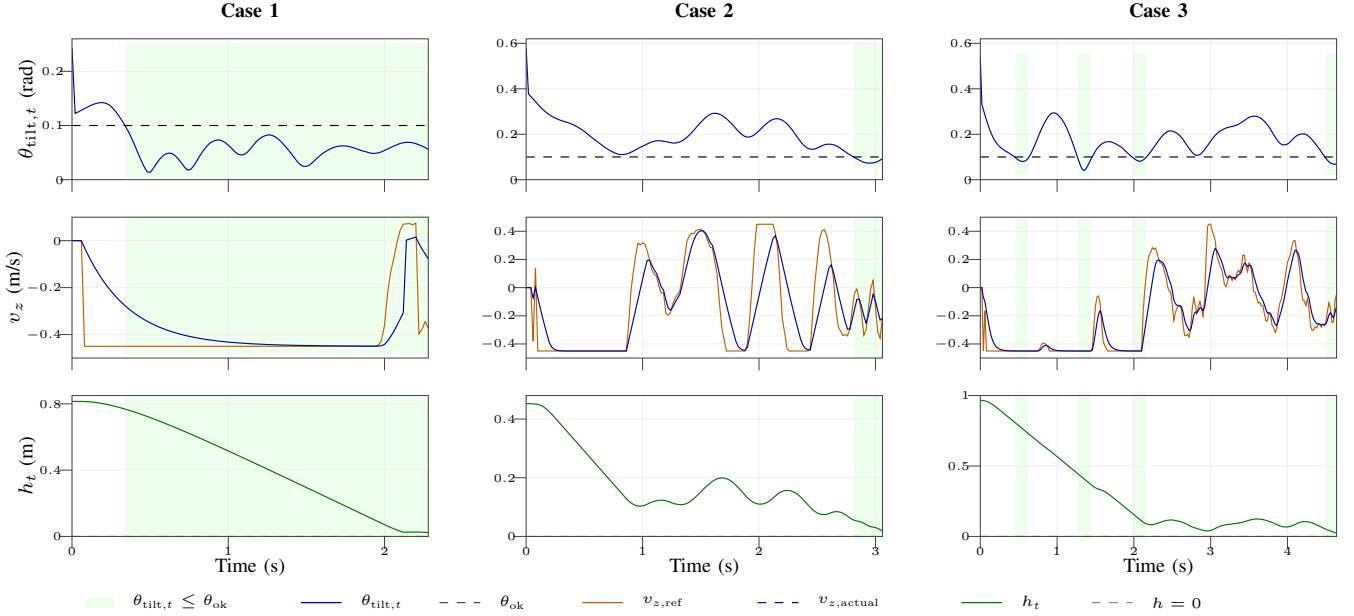}}
    \caption{Policy behavior under wave disturbances. Shaded regions denote safe touchdown intervals with  $\theta_{\mathrm{tilt},t} \leq \theta_{\mathrm{ok}}$.}
    \label{fig:isaacsim_policy_phase}
\end{figure*}

\section{Experiments and Analysis}
\label{sec:experiments}

This section evaluates WaveLander through policy behavior analysis, quantitative
baseline comparison, SITL transfer, and a representative real-world test.
\subsection{Policy Behavior Analysis}
\label{subsec:policy_behavior}

We analyze the learned high-level policy under selected wave-disturbed platform motions. For visualization, safe touchdown intervals are defined by
\(\theta_{\mathrm{tilt},t}\le\theta_{\mathrm{ok}}\), where
\(\theta_{\mathrm{ok}}=0.10~\mathrm{rad}\) is the same touchdown attitude threshold used in the quantitative evaluation. 
Fig.~\ref{fig:isaacsim_policy_phase} shows three representative cases. In
Case~1, the platform remains within the safe tilt bound for most of the
rollout, and the policy commands a smooth descent to touchdown. In Case~2, the
UAV starts close to the platform under an unfavorable attitude; the policy
therefore holds or retreats before selecting a suitable safe window to land. In
Case~3, stronger platform motion produces short safe windows. The policy mainly
continues descent and commands retreat-like motion only when the UAV is close to
the platform under unfavorable tilt, then lands when a safe window appears.

These results indicate that the learned policy is not a fixed descent law. It
adapts the vertical command according to relative height, vertical velocity,
platform attitude, and tilt rate. The actor uses a GRU-based policy, which retains
a short-term memory of past observations. Nevertheless, the observed timing
behavior is learned from the reward design rather than from an explicit
predictive model of future platform motion.

\subsection{Experimental Setup}
\label{subsec:experimental_setup}

The complete framework is evaluated in three settings:
\begin{itemize}
    \item \textbf{MuJoCo simulation:} quantitative baseline comparison under
    randomized wave-disturbed platform motions.
    \item \textbf{Isaac Sim SITL:} transfer to a high-fidelity simulation stack
    using Pegasus~\cite{jacinto2024pegasus} and an ArduPilot-based flight controller.
    \item \textbf{Real-world deployment:} a representative deployment-oriented
    test on a physical UAV system.
\end{itemize}
ROS~2~\cite{macenski2022ros2} is used for SITL and deployment integration. All methods share the same hierarchical structure: a low-level flight
controller handles attitude stabilization and lateral tracking, while the
evaluated strategy provides the vertical velocity command. 

\subsection{MuJoCo Evaluation and Baseline Comparison}
\label{subsec:mujoco_results}

The quantitative evaluation is conducted in the wave-disturbed MuJoCo
environments~\cite{todorov2012mujoco}. A video of the MuJoCo experiments is available online.\footnote{
 Supplementary video: \url{https://youtu.be/XaS_Jkk0rGU} } The platform uses the same randomized
motion model described in Section~\ref{subsec:wave_platform}, where heave, roll,
and pitch are generated by a spectral sum of sinusoidal components with
Ornstein--Uhlenbeck noise and exponential smoothing. In this evaluation, six
sinusoidal components are used, with frequencies uniformly sampled from
$0.05$ to $1.0~\mathrm{Hz}$.

WaveLander is compared with a constant-descent baseline that commands a fixed
vertical velocity of $-0.3~\mathrm{m/s}$ after lateral alignment. Both methods
use the same lateral controller, so the comparison isolates the vertical landing
strategy. Let $\theta_{\rm td}$ denote the touchdown attitude mismatch. A
touchdown is counted as successful when contact occurs with
$\theta_{\rm td}\leq0.1~\mathrm{rad}$.
Three platform-motion cases are evaluated, with
maximum roll/pitch motion ranges of $30^\circ$, $40^\circ$, and $60^\circ$.
WaveLander is evaluated on two UAV configurations, denoted
CF2 and X2. These configurations use the same high-level
landing policy interface but differ in vehicle dynamics and
low-level flight-control characteristics. The platform-motion
labels 30$^\circ$, 40$^\circ$, and 60$^\circ$ denote the
maximum roll and pitch range applied independently to the
moving platform. Therefore, these values should be interpreted
as per-axis attitude bounds rather than the maximum resultant
tilt angle. When roll and pitch vary simultaneously, the
resultant platform tilt can exceed the nominal per-axis bound.
For each case, 100 paired random seeds are used for both
methods.

\begin{table}[htbp]
\centering
\caption{Percentage of landings with touchdown attitude mismatch $\le 0.10$ rad in MuJoCo under different platform motion bounds.}
\label{tab:mujoco_mismatch_0.1}
\renewcommand{\arraystretch}{1.1}
\setlength{\tabcolsep}{6pt}
\begin{tabular}{lccc}
\hline
\textbf{Method} & \textbf{30$^\circ$} & \textbf{40$^\circ$} & \textbf{60$^\circ$} \\
\hline
Constant descent (CF2) & 33\% & 14\% &  8\% \\
Constant descent (X2)  & 27\% & 21\% &  5\% \\
WaveLander (CF2)       & 66\% & 60\% & 39\% \\
WaveLander (X2)        & 76\% & 54\% & 30\% \\
\hline
\end{tabular}
\vspace{0.5mm}

\footnotesize A touchdown with $\theta_{\mathrm{td}}\le 0.10~\mathrm{rad}$ indicates a high-quality landing with minimal platform tilt at contact.
\end{table}

As shown in Table~\ref{tab:mujoco_mismatch_0.1}, WaveLander
consistently improves the strict touchdown success rate under
the criterion $\theta_{\rm td}\leq 0.10~\mathrm{rad}$. For CF2,
the success rate increases from 33\%, 14\%, and 8\% to
66\%, 60\%, and 39\% across the $30^\circ$, $40^\circ$, and
$60^\circ$ platform-motion cases, respectively. For X2, the
corresponding improvement is from 27\%, 21\%, and 5\% to
76\%, 54\%, and 30\%. These results indicate that the learned
vertical policy improves touchdown timing relative to the
constant-descent baseline, especially under the $30^\circ$ and
$40^\circ$ motion cases.

\begin{figure}[htbp]
    \centering
    \input{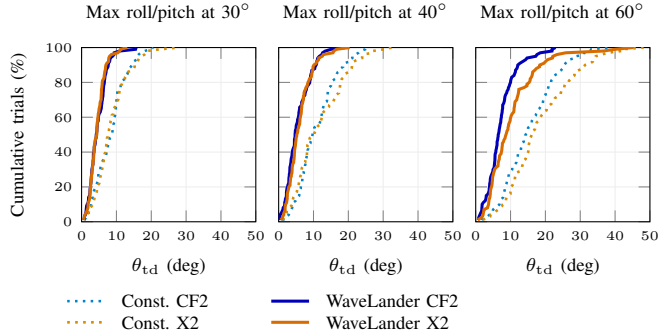}
    \caption{Cumulative distribution of touchdown attitude mismatch
$\theta_{\rm td}$ in MuJoCo. WaveLander shifts the distribution toward
lower mismatch values, indicating safer touchdown timing under
wave-induced platform motion.}
    \label{fig:mujoco_touchdown_error_cdf}
\end{figure}

The CDFs in Fig.~\ref{fig:mujoco_touchdown_error_cdf} further support this interpretation. In the
30$^\circ$ and 40$^\circ$ cases, the WaveLander curves are shifted
toward smaller touchdown mismatch values compared with the
constant-descent baseline, indicating that more trials terminate
with favorable platform attitude at contact. In the 60$^\circ$
stress-test case, the gap narrows because the platform motion is
more aggressive and favorable touchdown windows become shorter
and less frequent. Even in this case, however, WaveLander retains
a higher success rate, suggesting that the learned timing strategy
provides improved robustness under wave-induced attitude motion.

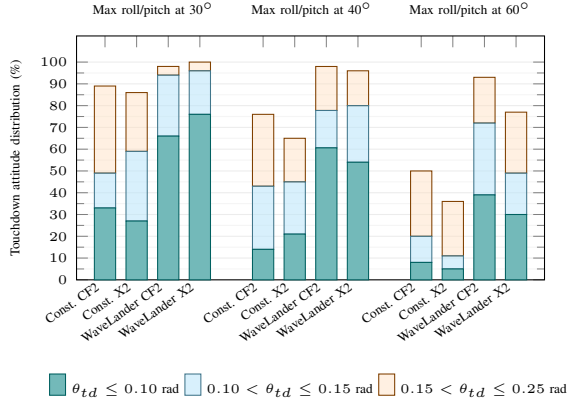
\begin{figure}[htbp]
    \centering

\begin{tikzpicture}
\begin{axis}[
  ybar stacked,
  bar width=8pt,
  width=0.90\linewidth,
  height=0.20\textheight,
  ymin=0,
  ymax=112,
  ylabel={Touchdown attitude distribution (\%)},
  xtick={1,2,3,4,6,7,8,9,11,12,13,14},
  xticklabels={
    Const. CF2, Const. X2, WaveLander CF2, WaveLander X2,
    Const. CF2, Const. X2, WaveLander CF2, WaveLander X2,
    Const. CF2, Const. X2, WaveLander CF2, WaveLander X2
  },
  x tick label style={rotate=35, anchor=east, font=\tiny},
  xmin=0.1,
  xmax=14.9,
  enlarge x limits=false,
  clip=false,
  ymajorgrids=true,
  yminorgrids=true,
  ytick={0,10,20,30,40,50,60,70,80,90,100},
  minor y tick num=1,
  grid style={draw=gray!15},
  minor grid style={draw=gray!8},
  ticklabel style={font=\tiny},
  label style={font=\tiny},
  legend style={
    font=\tiny,
    draw=none,
    fill=none,
    at={(0.5,-0.36)},
    anchor=north,
    legend columns=3
  },
  legend cell align=left,
  legend entries={
    $\theta_{td} \leq 0.10$ rad,
    $0.10 < \theta_{td} \leq 0.15$ rad,
    $0.15 < \theta_{td} \leq 0.25$ rad
  }
]

\addplot[
  fill=teal!55,
  draw=teal!75!black,
  fill opacity=1
] coordinates {
  (1,33.0) (2,27.0) (3,66.0) (4,76.0)
  (6,14.0) (7,21.0) (8,60.60606060606061) (9,54.0)
  (11,8.0) (12,5.0) (13,39.0) (14,30.0)
};

\addplot[
  fill=cyan!25,
  draw=cyan!45!black,
  fill opacity=0.55
] coordinates {
  (1,16.0) (2,32.0) (3,28.0) (4,20.0)
  (6,29.0) (7,24.0) (8,17.171717171717162) (9,26.0)
  (11,12.0) (12,6.0) (13,33.0) (14,19.0)
};

\addplot[
  fill=orange!25,
  draw=orange!50!black,
  fill opacity=0.45
] coordinates {
  (1,40.0) (2,27.0) (3,4.0) (4,4.0)
  (6,33.0) (7,20.0) (8,20.202020202020208) (9,16.0)
  (11,30.0) (12,25.0) (13,21.0) (14,28.0)
};

\node[font=\tiny, anchor=south] at (rel axis cs:0.164,1.04) {Max roll/pitch at 30$^\circ$};
\node[font=\tiny, anchor=south] at (rel axis cs:0.503,1.04) {Max roll/pitch at 40$^\circ$};
\node[font=\tiny, anchor=south] at (rel axis cs:0.841,1.04) {Max roll/pitch at 60$^\circ$};

\end{axis}
\end{tikzpicture}
    \caption{Distribution of touchdown attitude mismatch in MuJoCo under different per-axis platform-motion bounds. The stacked bars show the percentage of trials falling within each \(\theta_{\mathrm{td}}\) range. Only the lowest bin, \(\theta_{\mathrm{td}}\le 0.10~\mathrm{rad}\), is used as the strict success criterion in Table~\ref{tab:mujoco_mismatch_0.1}.}
    \label{fig:mujoco_safe_rate}
\end{figure}

\subsection{Transfer to Isaac Sim with SITL}
\label{subsec:sitl}

To evaluate deployment compatibility, the trained policy is transferred to
NVIDIA Isaac Sim using Pegasus~\cite{jacinto2024pegasus} and integrated with
an ArduPilot-based flight controller in a software-in-the-loop configuration.
ROS~2 connects the simulator, policy inference module, and flight-control
interface~\cite{macenski2022ros2}.

The policy is deployed without additional training or parameter tuning. The
same four-dimensional observation
$o_t=[h,\dot h,\theta_{tilt},\dot{\theta}_{tilt}]^\top$ and scalar vertical
velocity command are used. This experiment verifies that the learned high-level
command can be executed in an autopilot-integrated simulation stack without
changing the policy interface.

\subsection{Real-World Deployment}
\label{subsec:real_world_deployment}

WaveLander is further prepared for deployment on a physical UAV system using
the same hierarchical structure as in simulation. The flight controller handles
stabilization and lateral tracking, while the learned policy provides the
vertical velocity reference.

A representative deployment-oriented test was conducted under easterly wind,
with an average wind speed of approximately \(10~\mathrm{kt}\) and gusts up to
\(18~\mathrm{kt}\). The observed wave height was approximately
\(0.8\)--\(0.9~\mathrm{m}\), with a period of about \(5~\mathrm{s}\).

\begin{figure}[htbp]
    \centering
    \includegraphics[width=\columnwidth]{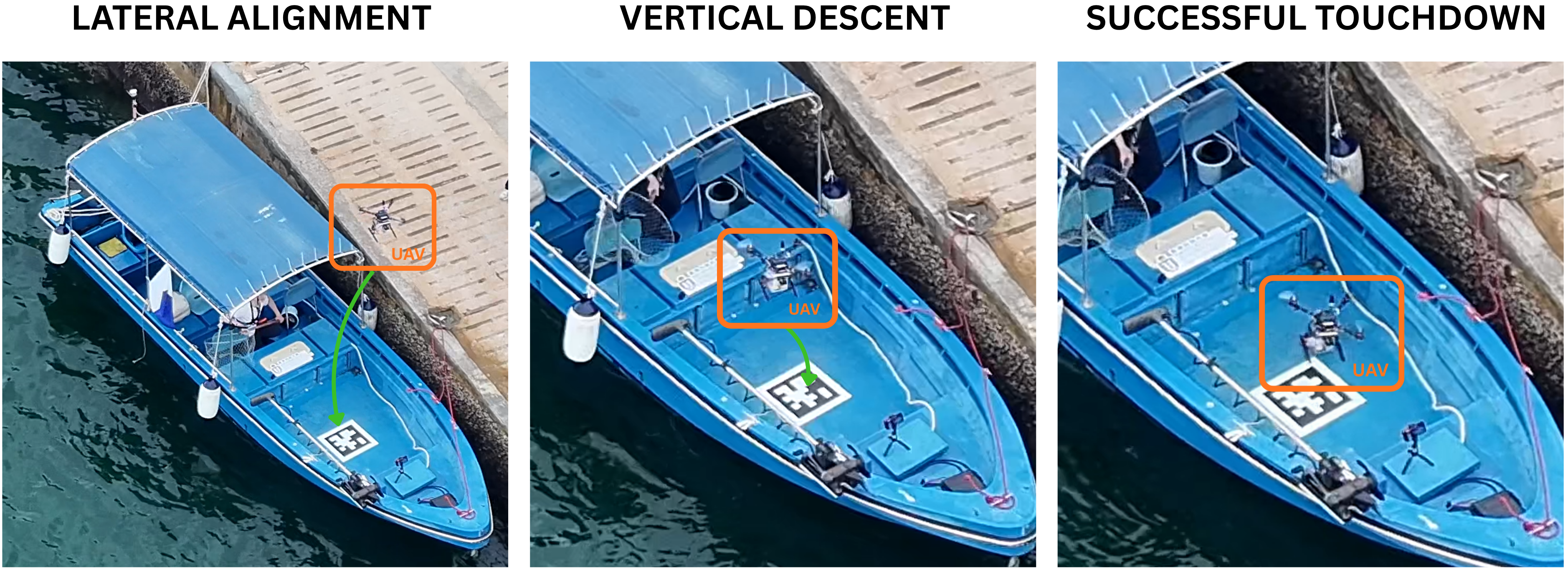}
    \caption{Real-world deployment setup under moderate wind and wave
    conditions.}
    \label{fig:real_world_setup}
\end{figure}

Fig.~\ref{fig:real_world_setup} shows the deployment sequence, including
lateral alignment, vertical descent, and touchdown under the tested condition.
This experiment demonstrates the feasibility of implementing the compact
platform-relative observation and vertical velocity command interface on a
physical UAV system. Large-scale real-world validation under broader sea states
is left for future work.

\subsection{Discussion}
\label{subsec:discussion}

The experiments provide complementary evidence for the effectiveness and
deployability of WaveLander. The policy-behavior analysis shows that the
learned actor is not a fixed descent law: it descends smoothly under mild
platform motion, but slows, holds, or retreats when the platform attitude
becomes unfavorable. The MuJoCo comparison is consistent with this interpretation: WaveLander produces a higher fraction of low-attitude-mismatch
touchdowns than the constant-descent baseline, indicating that the
main improvement comes from more selective touchdown timing rather
than from executing a fixed descent command.

These results indicate that WaveLander improves landing performance mainly by
learning touchdown timing. Since the lateral controller is shared across
methods, the MuJoCo improvement can be attributed primarily to the learned
vertical decision policy rather than to lateral tracking differences. Using only
relative height, vertical velocity, platform attitude, and short-term tilt-rate
information, the policy learns to exploit favorable touchdown phases without an
explicit switching rule or future platform-motion predictor.

The hierarchical design is also important for deployment. By restricting the
learned component to a scalar vertical velocity reference, WaveLander remains
compatible with conventional low-level flight controllers and reduces the
sim-to-real burden. The same compact interface can therefore be inserted as a
high-level module above existing stabilization and lateral tracking loops.

Several limitations define the current scope of the study. The wave-platform
model is intentionally lightweight and is mainly used to generate repeatable
heave--tilt disturbances for policy evaluation, rather than to reproduce full
marine hydrodynamics. Similarly, near-surface aerodynamic effects, sensing
noise, and latency are not modeled in detail, but the hierarchical design and
low-speed vertical command help reduce the sensitivity to these effects. Although
the actor is recurrent and observes short-term tilt variation, it does not
include an explicit future platform-motion prediction module; instead, the
touchdown timing behavior is learned implicitly from the observation history and
reward structure. Finally, the real-world experiment is intended as a
deployment-oriented validation of the interface and control stack, while a
larger-scale field evaluation is left for future work. Future extensions will
consider perception uncertainty, richer temporal modeling, and higher-fidelity
platform and aerodynamic effects.

\section{Conclusion}
\label{sec:conclusion}

This paper presented \textit{WaveLander}, a hierarchical control framework via RL for UAV landing on wave-disturbed platforms. By restricting the
learned component to a scalar vertical velocity reference and relying on a
conventional low-level flight controller for stabilization and lateral tracking,
the proposed framework reduces dynamic platform landing to a compact
observation-based vertical decision task. The policy is trained under
randomized wave-induced platform motions and operates without explicit
threshold-based switching during deployment. Simulation results show that the
proposed approach adjusts vertical landing behavior according to relative
height and platform attitude, improving robustness over the constant-descent baseline. Future work will focus on perception uncertainty,
higher-fidelity aerodynamic effects, temporal observation augmentation, and
large-scale real-world validation.

{\small
\bibliographystyle{ieeetr}
\bibliography{reference}

@article{gupta2023landing,
  author  = {Gupta, Parakh M. and Pairet, {\'E}ric and Nascimento, Tiago and Saska, Martin},
  title   = {Landing a {UAV} in Harsh Winds and Turbulent Open Waters},
  journal = {IEEE Robotics and Automation Letters},
  volume  = {8},
  number  = {2},
  pages   = {744--751},
  year    = {2023}
}

@article{rodriguez2019deep,
  author  = {Rodriguez-Ramos, Alejandro and Sampedro, Carlos and Bavle, Hriday and de la Puente, Paloma and Campoy, Pascual},
  title   = {A Deep Reinforcement Learning Strategy for {UAV} Autonomous Landing on a Moving Platform},
  journal = {Journal of Intelligent \& Robotic Systems},
  volume  = {93},
  pages   = {351--366},
  year    = {2019}
}

@article{goldschmid2024rl,
  author  = {Goldschmid, Pascal and Ahmad, Aamir},
  title   = {Reinforcement Learning Based Autonomous Multi-Rotor Landing on Moving Platforms},
  journal = {Autonomous Robots},
  volume  = {48},
  number  = {4},
  pages   = {13},
  year    = {2024}
}

@article{saj2022robust,
  author  = {Saj, Vishnu and Lee, Bochan and Kalathil, Dileep and Benedict, Moble},
  title   = {Robust Reinforcement Learning Algorithm for Vision-Based Ship Landing of {UAV}s},
  journal = {arXiv preprint arXiv:2209.08381},
  year    = {2022}
}

@article{peter2024landerai,
  author  = {Peter, Robinroy and Ratnabala, Lavanya and Aschu, Demetros and Fedoseev, Aleksey and Tsetserukou, Dzmitry},
  title   = {Lander.AI: Adaptive Landing Behavior Agent for Expertise in 3D Dynamic Platform Landings},
  journal = {arXiv preprint arXiv:2403.06572},
  year    = {2024}
}

@inproceedings{todorov2012mujoco,
  title={{MuJoCo}: A physics engine for model-based control},
  author={Todorov, Emanuel and Erez, Tom and Tassa, Yuval},
  booktitle={2012 IEEE/RSJ International Conference on Intelligent Robots and Systems},
  pages={5026--5033},
  year={2012}
}

@article{macenski2022ros2,
  title={{Robot Operating System} 2: Design, architecture, and uses in the wild},
  author={Macenski, Steven and others},
  journal={Science Robotics},
  volume={7},
  number={66},
  pages={eabm6074},
  year={2022}
}

@article{mittal2025isaaclab,
  title={{Isaac Lab}: A GPU-Accelerated Simulation Framework for Multi-Modal Robot Learning},
  author={Mittal, Mayank and Roth, Pascal and Tigue, James and Richard, Antoine and Zhang, Octi and Du, Peter and Serrano-Mu{\~n}oz, Antonio and Yao, Xinjie and Zurbr{\"u}gg, Ren{\'e} and Rudin, Nikita and others},
  journal={arXiv preprint arXiv:2511.04831},
  year={2025}
}

@article{schwarke2025rslrl,
  title={{RSL-RL}: A Learning Library for Robotics Research},
  author={Schwarke, Clemens and Mittal, Mayank and Rudin, Nikita and Hoeller, David and Hutter, Marco},
  journal={arXiv preprint arXiv:2509.10771},
  year={2025},
  doi={10.48550/arXiv.2509.10771}
}

@article{uhlenbeck1930theory,
  author  = {Uhlenbeck, G. E. and Ornstein, L. S.},
  title   = {On the Theory of the Brownian Motion},
  journal = {Physical Review},
  volume  = {36},
  number  = {5},
  pages   = {823--841},
  year    = {1930}
}

@article{schulman2017ppo,
  title   = {Proximal Policy Optimization Algorithms},
  author  = {Schulman, John and Wolski, Filip and Dhariwal, Prafulla and Radford, Alec and Klimov, Oleg},
  journal = {arXiv preprint arXiv:1707.06347},
  year    = {2017}
}

@article{wang2022quadrotor,
  title={Quadrotor autonomous landing on moving platform},
  author={Wang, Pengyu and Wang, Chaoqun and Wang, Jiankun and Meng, Max Q-H},
  journal={Procedia Computer Science},
  volume={209},
  pages={40--49},
  year={2022},
  publisher={Elsevier}
}

@inproceedings{lin2024coastal,
  title={Coastal Underwater Evidence Search System with Surface-Underwater Collaboration},
  author={Lin, Hin Wang and Wang, Pengyu and Yang, Zhaohua and Leung, Ka Chun and Bao, Fangming and Kui, Ka Yu and Xu, Jian Xiang Erik and Shi, Ling},
  booktitle={IEEE International Conference on Control, Automation, Robotics and Vision},
  pages={1047--1053},
  year={2024}
}

@article{zhang2025aerial,
  title={Aerial-marine cross-domain uncrewed systems: An overview of cyberphysical coordination frameworks for marine applications},
  author={Zhang, Hai-Tao and Hu, Bin-Bin and Liu, Bin and Ding, Jianing and Zhao, Jin and Su, Housheng and Zhang, Yunfei and Zhu, Cheng and Yuan, Ye and Shi, Yang},
  journal={IEEE Control Systems},
  volume={45},
  number={4},
  pages={28--45},
  year={2025},
  publisher={IEEE}
}

@article{li2024separation,
  title={A Separation and Rendezvous Control Method for the {UAV-USV} System Based on Distributed {NMPC}},
  author={Li, Shilong and Zhu, Yakun and Guo, Ge and Yuan, Pengfei and Bai, Jianguo},
  journal={IEEE Transactions on Intelligent Vehicles},
  year={2024},
  volume={9},
  number={11},
  pages={7251-7263},
  publisher={IEEE}
}

@inproceedings{polvara2018toward,
  title={Toward end-to-end control for {UAV} autonomous landing via deep reinforcement learning},
  author={Polvara, Riccardo and Patacchiola, Massimiliano and Sharma, Sanjay and Wan, Jian and Manning, Andrew and Sutton, Robert and Cangelosi, Angelo},
  booktitle={2018 International Conference on Unmanned Aircraft Systems},
  pages={115--123},
  year={2018}
}

@article{li2022synchronized,
  title={Synchronized motion-Based {UAV-USV} cooperative autonomous landing},
  author={Li, Wenzhan and Ge, Yuan and Guan, Zhihong and Ye, Gang},
  journal={Journal of Marine Science and Engineering},
  volume={10},
  number={9},
  pages={1214},
  year={2022}
}

@article{li2023nmpc,
  title={{NMPC}-based {UAV-USV} cooperative tracking and landing},
  author={Li, Wenzhan and Ge, Yuan and Guan, Zhihong and Gao, Hongbo and Feng, Haoyu},
  journal={Journal of the Franklin Institute},
  volume={360},
  number={11},
  pages={7481--7500},
  year={2023},
  publisher={Elsevier}
}

@article{xu2024manipulator,
  title={A manipulator-assisted multiple {UAV} landing system for {USV} subject to disturbance},
  author={Xu, Ruoyu and Liu, Chongfeng and Cao, Zhongzhong and Wang, Yuquan and Qian, Huihuan},
  journal={Ocean Engineering},
  volume={299},
  pages={117306},
  year={2024},
  publisher={Elsevier}
}

@book{sutton2018reinforcement,
  author    = {Sutton, Richard S. and Barto, Andrew G.},
  title     = {Reinforcement Learning: An Introduction},
  edition   = {2},
  publisher = {MIT Press},
  address   = {Cambridge, MA, USA},
  year      = {2018}
}

@inproceedings{jacinto2024pegasus,
  title={Pegasus simulator: An isaac sim framework for multiple aerial vehicles simulation},
  author={Jacinto, Marcelo and Pinto, Jo{\~a}o and Patrikar, Jay and Keller, John and Cunha, Rita and Scherer, Sebastian and Pascoal, Ant{\'o}nio},
  booktitle={2024 International Conference on Unmanned Aircraft Systems},
  pages={917--922},
  year={2024}
}
}

\end{document}